# Multiple Images Recovery Using a Single Affine Transformation

Bo Jiang[1], Chris Ding[2,1] and Bin Luo[1]

[1]School of Computer Science and Technology, Anhui University, Hefei, 230601, China
[2]CSE Department, University of Texas at Arlington, Arlington, TX 76019, USA
jiangbo@ahu.edu.cn, chqding@uta.edu, luobin@ahu.edu.cn

## Abstract

In many real-world applications, image data often come with noises, corruptions or large errors. One approach to deal with noise image data is to use data recovery techniques which aim to recover the true uncorrupted signals from the observed noise images. In this paper, we first introduce a novel *corruption recovery transformation* (CRT) model which aims to recover multiple (or a collection of) corrupted images using a single affine transformation. Then, we show that the introduced CRT can be efficiently constructed through learning from training data. Once CRT is learned, we can recover the true signals from the new incoming/test corrupted images explicitly. As an application, we apply our CRT to image recognition task. Experimental results on six image datasets demonstrate that the proposed CRT model is effective in recovering noise image data and thus leads to better recognition results.

## 1 Introduction

In many real-world applications, image data often come with various kinds of noises and errors due to different reasons, for example images are corrupted, data components are missing, errors due to human recording or machine malfunction, etc. These noises and corruptions usually make the recognition/learning task of image data more challengeable and significantly affect the recognition results. One popular approach to deal with noise image data is to use data recovery/reconstruction techniques which aim to recover the true uncorrupted signals from the observed noise data.

There is a very large number of studies on image data recovery or noise/corruption removal on images. Among them, one kind of widely used methods is to use low-rank matrix factorization technique, such as Principal Component Analysis (PCA) [Duda *et al.*, 2001], Nonnegative Matrix Factorization (NMF)[Seung and Lee, 2001; Cai and He, 2011] and Dictionary Learning [Aharon *et al.*, 2006], etc. These methods can effectively recover the true signals in low-rank space from images with Gaussian type noise. However, sometimes the noises are large such as outliers, corrupted images, different illuminations, etc. For these large noises or gross errors, traditional matrix factorization methods usually break down. Recent works use more robust matrix norms such as $\ell_1$ norm [Ke and Kanade, 2005; Kasiviswanathan *et al.*, 2012; Peng *et al.*, 2010; Zhao and Cham, 2011; Zhang *et al.*, 2011], $\ell_{21}$ norm [Ding *et al.*, 2006; Kwak, 2008; Kong *et al.*, 2011] to develop robust matrix factorization formulations which have been shown robustly in dealing with gross errors or outliers. In additional to matrix factorization, rank regularization approaches have also been applied to reduce the rank of the data inexplicitly and thus conduct recovery task effectively, even in the presence of noise [Cai *et al.*, 2010; Ma *et al.*, 2009; Liu *et al.*, 2010; Liu and Yan, 2011]. These methods generally use the nuclear norm as the main component for rank reduction [Fazel, 2002; Recht *et al.*, 2010]. One main advantage of these nuclear norm approaches is that the nuclear-norm is the convex envelop of matrix rank and thus the optimization is usually convex. A unique optimal solution usually exists. In the $\ell_1$ norm based approach [Wright *et al.*, NIPS 2009; Chandrasekaran *et al.*, 2009], authors have shown the good effects for recovering true signals from large corruptions.

In this paper, we first introduce a novel *corruption recovery transformation* (CRT). This transformation aims to recover multiple (or a collection of) corrupted images explicitly using a single affine transformation. To the best of our knowledge, this problem has not been done or emphasized before. Then, we show that the introduced CRT can be efficiently constructed through learning from training data, and propose a robust CRT learning model and algorithm. Once our CRT is learned, we can explicitly recover the true signals for the new incoming/test corrupted images effectively. As an application, we apply our CRT to image recognition task and propose a new image classification method. Experimental results on six image datasets demonstrate that the proposed CRT model is effective in recovering noise image data which obviously improves the recognition results.

## 2 Problem Statement

In this paper, we aim to show how to use a single affine transformation to recover multiple corrupted images. To the best of our knowledge, this problem has not been done or emphasized before. Formally, this paper addresses the following problem.

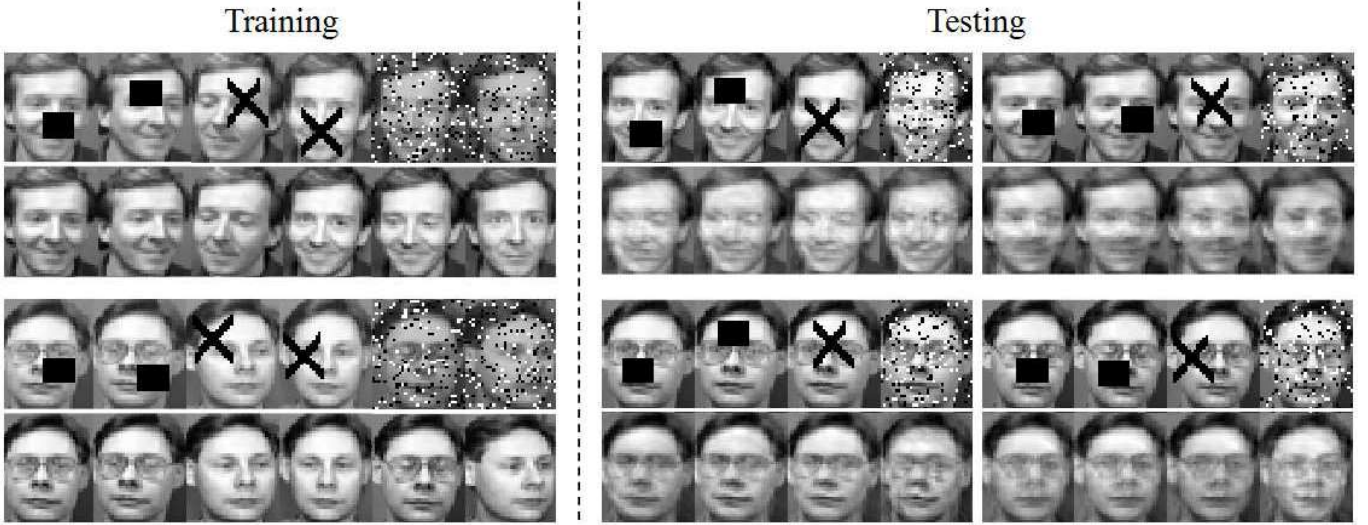

Figure 1: LEFT:Examples of training images for CRT (in each panel, top row shows corrupted images; bottom row shows ground-truth images). RIGHT: Image recovery results using CRT (in each panel, top row shows the multiple corrupted images; bottom row shows the recovery results). Note that, the corrupted test images are not included in the training set and all recoveries use a single CRT.

## 2.1 Corruption recovery transformation

Let $X = (x_1, x_2, \cdots, x_n) \in \mathbb{R}^{p \times n}$ be the collection of corrupted images. We write $X = X^0 + E$, where $X^0 = (x_1^0, x_2^0, \cdots x_n^0)$ is the true (uncorrupted) data and $E$ is the noise or corruption. Let $A \in \mathbb{R}^{p \times p}$ be the affine transformation (we will explain how it is constructed later). $A$ has the ability that can automatically recover $X^0$ from $X$ as,

$$X^0 \simeq AX \quad \text{or} \quad x_i^0 \simeq Ax_i, \quad i = 1...n \qquad (1)$$

We call $A$ as Corruption Recovery Transformation (CRT). The main contribution in this paper is to show that this CRT does exist and it can be efficiently constructed through learning from training data. In the following, we first show the recovery ability of the proposed (learned) CRT on some data, and then explain CRT in detail in Section 3.

## 2.2 Illustration

Figure 1 shows the training images and recovered images on AT&T face dataset. The dataset contains 40 persons with 10 slightly different images for each person. We randomly select 8 images of each person to train the CRT, and use the rest 2 images for testing. On all images, we add three kinds of corruption noises including 'block', 'cross' and 'pepper & salt' as shown in Figure 1. Figure 1 (LEFT) shows examples of training images which are used to learn the CRT. Each panel shows examples of the corrupted images and corresponding ground-truth images. Details for learning CRT is explained in the following section.

Figure 1 (RIGHT) shows examples of CRT recovery results. Each panel shows a test image with 4 different corruptions, and the recovered images. Note that, these test corrupted images are not included in training set and all recoveries use the same single CRT. Here, we can observe,

- CRT (linear transformation) can reasonably remove severe corruptions (which are normally believed to require nonlinear process).
- CRT can recover different type of corruptions, such as 'block', 'cross' and 'pepper & salt'.
- The quality of CRT recovery is quite reasonable, considering the simplicity of this approach.

The rest of this paper will focus on the efficient construction of the CRT and experimental validation.

## 3 Learning CRT

In this section, we demonstrate how to construct CRT through learning from training images. Formally, let $Z = (z_1, \cdots z_m) \in \mathbb{R}^{p \times m}$ be the input corrupted data (image), and $Z^0 = (z_1^0, \cdots z_m^0)$ be corresponding ground-truth (uncorrupted) data. Here $p = hw$, $h$ and $w$ are the hight and width of each image, and all images have been stored as vectors as in the standard PCA of images [Duda *et al.*, 2001].

Note that ground-truth (uncorrupted) data can be obtained in several ways. In some situations, we do not know the ground-truth clear images, and we use some low-rank recovery methods, such RPCA [Wright *et al.*, NIPS 2009], LRR [Liu *et al.*, 2010], to learn an approximation of the clean data — this is a kind of preprocessing. In other cases, the input data have small Gaussian noises and but no big corruptions. We leave them untouched and regard them as the "clean" data. We generate some artificial corruptions to the data and use this corrupted data as the noisy data. In both situations, the clean data and noisy data are obtained before the learning of the transformation $A$.

We can simply learn CRT $A$ by solving the following prob-lem,
$$\min_A \sum_{i=1}^m \|z_i^0 - Az_i\|_2^2 = \|Z^0 - AZ\|_F^2. \quad (2)$$

It is well known that the optimal solution $A^* = Z^0 Z^T (ZZ^T)^{-1}$.

Once we have learned the optimal $A^*$, for any new incoming test corrupted images $x_i \in X$ (see Eq.(1)), we recover its true signal data $x_i^0$ as

$$x_i^0 \simeq A^* x_i \quad (3)$$

### 3.1 Robust CRT

To improve the robustness of CRT, we first incorporate regularization on the affine transformation $A$ to achieve smoothness, low-rank, etc, i.e.,

$$\min_A \|Z^0 - AZ\|_F^2 + \lambda \mathrm{reg}(A), \quad (4)$$

where $\lambda > 0$ is a parameter and $\mathrm{reg}(A)$ is a regularization term of $A$. There are many different regularization ways such as nuclear (trace) norm, $\ell_2$ norm and total variation, etc. One popular way is to use nuclear norm to induce low-rank constraint. In this paper we use nuclear norm regularization and thus optimize the following problem,

$$\min_A \|Z^0 - AZ\|_F^2 + \lambda \|A\|_*, \quad (5)$$

where $\|A\|_* = \mathrm{Tr}(AA^T)^{1/2}$ is nuclear norm.

Second, since $\ell_{2,1}$ norm is more robust to outliers, thus it is more reasonable to use $\ell_{2,1}$ norm instead of $\ell_2$ norm function to achieve robustness of CRT. Thus, our robust CRT can finally be formulated as,

$$\min_A \|Z^0 - AZ\|_{2,1} + \lambda \|A\|_* \quad (6)$$

where $\|M\|_{2,1} = \sum_j (\sum_i |M_{ij}|^2)^{1/2}$ is $\ell_{2,1}$ norm[1].

This problem formulation is convex. We will present an effective algorithm to obtain the global optimal solution for this problem, as shown in section 3.5 in detail.

### 3.2 Comparison of RPCA, LRR and CRT

From model formulation aspect, our CRT has some resemblance to previous well known RPCA [Wright *et al.*, NIPS 2009] and LRR [Liu *et al.*, 2010] methods. However, CRT differs fundamentally from previous approaches in the following. RPCA [Wright *et al.*, NIPS 2009] aims to find a low-rank approximation $Y$ for input data $X$ by solving,

$$\min_Y \|X - Y\|_1 + \lambda \|Y\|_* \quad (7)$$

LRR [Liu *et al.*, 2010] aims to find the low-rank representation $Y$ for a collection of data vectors $X$ by solving,

$$\min_Y \|X - DY\|_{2,1} + \lambda \|Y\|_* \quad (8)$$

where $D$ is the dictionary. LRR can also be used to learn a low-rank affinity graph for data points by setting $D = X$.

Differently, in our CRT (Eq.(6)), it aims to learn an affine transformation $A$ that can recover true (uncorrupted) data $Z^0$ from corrupted data $Z$. Thus, our CRT can be applied in semi-supervise/supervise learning tasks. In the following, we first discuss two important issues relating to our robust CRT and then present an effective algorithm to solve it.

---
[1]Here, one can use some other robust norms such as $\ell_{1,1}$ norm

### 3.3 Ground-truth training image $Z^0$

In some applications, the ground truth (uncorrupted) data $Z^0$ are hard to obtain. In these cases, we can obtain $Z^0$ approximately from $Z$ by using some robust recovery methods such as Robust Principal Component Analysis (RPCA) [Wright *et al.*, NIPS 2009], Low Rank Representation (LRR) [Liu *et al.*, 2010], etc. Figure 2 (LEFT) shows some examples on Yale-B face data (see Experiments in detail) where $Z^0$ are obtained using RPCA on original data $Z$. Using $\{Z^0, Z\}$, we can learn an optimal CRT $A^*$ using the above robust CRT (Eq.(6)). Then, we recover the test images using the learned $A^*$. Figure 2 (RIGHT) shows the test images and recovery results. We can note that the shading noise of test images can be suppressed in CRT recoveries.

### 3.4 Visualization of CRT

The computed CRT $A$ can be written as $A = (a_1, \cdots, a_p)$, where $a_j$ is the $j$-th column of $A$. We may consider these $a_j$'s as basis vectors of the affine transformation. Another equivalent way to see this is that for any test (corrupted) image $x \in \mathbb{R}^p$, we recover its true signal $x^0$ as

$$x^0 \simeq Ax = \sum_{k=1}^p x_k a_k \quad (9)$$

Thus, $(a_1, \cdots, a_p)$ are the basis of the CRT.

To show these basis vectors intuitively, we can display them as images (note that $a_j$ is a vector of length $p$). Figure 3 shows the first 32 columns images of $A^*$ due to limited space. Similar to eigenfaces [Turk and Pentland, 1991] and Laplacianfaces [He *et al.*, 2005], these basis vectors define a kind of robust space.

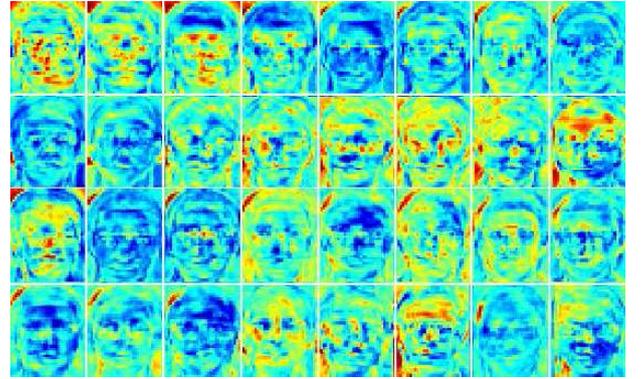

Figure 3: Basis vectors of CRT on AT&T face data

### 3.5 Computational algorithm

Our CRT learning problem Eq.(6) is a convex formulation. The global optimal solution can be efficiently computed using the following Augmented Lagrange Multiplier (ALM) algorithm.

We first convert problem Eq.(6) to the following equivalent problem,

$$\min_{A,F,E} \|E\|_{2,1} + \lambda \|F\|_* \quad (10)$$

$$s.t. \ F = A, E = Z^0 - AZ.$$

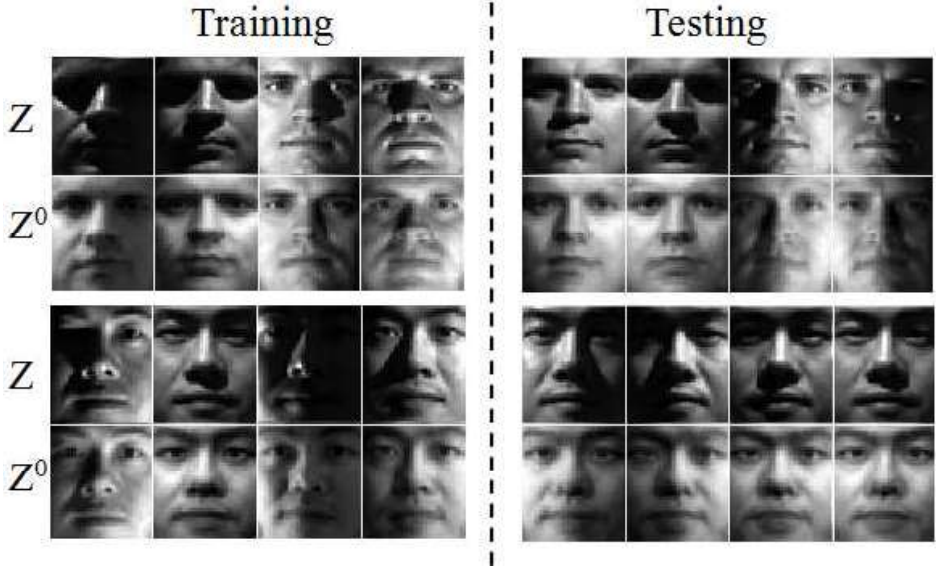

Figure 2: CRT recovery results on YaleB face data. LEFT: examples of original training images $Z$ with various shading noise and corresponding (approximate) true signals $Z^0$ computed by RPCA method. RIGHT: test images (1st row of each panel) and corresponding CRT recovery results (2nd row of each panel)

Then, ALM solves a sequence of sub-problems,

$$J = \|E\|_{2,1} + \langle \Lambda, A - F \rangle + \frac{\mu}{2}\|A - F\|_F^2 + \lambda\|F\|_*$$
$$+ \langle \Omega, Z^0 - AZ - E \rangle + \frac{\mu}{2}\|Z^0 - AZ - E\|_F^2. \quad (11)$$

where $\langle P, Q \rangle = \sum_{ij} P_{ij} Q_{ij} = \text{Tr} P^T Q$, $\Lambda, \Omega$ are Lagrange multipliers, $\mu$ is the penalty parameter. There are two main parts of the algorithm, i.e., solving the sub-problem and updating parameters.

**Step 1**. Solve $F$ while fixing $A, E$. That is

$$\min_F \lambda\|F\|_* + \frac{\mu}{2}\|F - (A + \frac{\Lambda}{\mu})\|_F^2. \quad (12)$$

It has the closed form solution using singular value thresholding method, i.e.,

$$\text{SVD}(A + \frac{\Lambda}{\mu}) = U\Sigma V^T, \quad F^* = U(\Sigma - \frac{\lambda}{\mu}I)_+ V^T, \quad (13)$$

where $\text{SVD}(M)$ is the singular value decomposition of $M$, and $(M_+)_{ij} = \frac{1}{2}(|M_{ij}| + M_{ij})$.

**Step 2**. Solve $E$ while fixing $A, F$. The problem becomes,

$$\min_E \|E\|_{2,1} + \frac{\mu}{2}\|E - (Z^0 - AZ + \frac{\Omega}{\mu})\|_F^2. \quad (14)$$

Let $P = Z^0 - AZ + \frac{\Omega}{\mu}$, then the optimal closed form solution for this problem is

$$E^*_{\cdot j} = \frac{P_{\cdot j}}{\|P_{\cdot j}\|_2} \max(\|P_{\cdot j}\|_2 - \frac{1}{\mu}, 0), \quad (15)$$

where $P_{\cdot j}, E^*_{\cdot j}$ are the $j$-th column of $P$ and $E^*$.

**Step 3**. Solve $A$ while fixing $E, F$. The problem becomes,

$$\min_A \langle \Lambda, A - F \rangle + \langle \Omega, Z^0 - AZ - E \rangle + \frac{\mu}{2}\|A - F\|_F^2$$
$$+ \frac{\mu}{2}\|Z^0 - AZ - E\|_F^2. \quad (16)$$

The optimal $A^*$ is obtained by setting derivative w.r.t. $A$ to zero, i.e.,

$$A^* = \left[F + (Z^0 - E + \frac{\Omega}{\mu})Z^T - \frac{\Lambda}{\mu}\right](I + ZZ^T)^{-1}. \quad (17)$$

**Step 4**. Update parameters $\Lambda, \Omega, \mu$ as

$$\Lambda \Leftarrow \Lambda + \mu(A - F) \quad (18)$$
$$\Omega \Leftarrow \Omega + \mu(Z^0 - AZ - E) \quad (19)$$
$$\mu \Leftarrow \rho\mu \quad (20)$$

where $\rho > 1$.

The algorithm iteratively conducts step 1-step4 until convergence. The complete algorithm is summarized in Algorithm 1.

## 4 Application: Image Classification

As an application of the proposed CRT, we applied it in image classification tasks. In short, the classification process has the following two main steps.

**Training step:** For training images $Z$, we first obtain their true signals $Z^0$ approximately using RPCA [Wright *et al.*, NIPS 2009] method. Then, we learn the optimal CRT $A^*$ from training set of images $\{Z^0, Z\}$ using the proposed Robust CRT (Eq.(6)).

**Algorithm 1** Robust CRT Algorithm

**Input:** Training data $Z, Z^0 \in \mathbb{R}^{p \times m}$, parameter $\lambda$
**Output:** The optimal CRT $A^*$
1: Initialize $\Lambda = 0, \Omega = 0, E = 0, \mu = 1e-6, \mu_{max} = 10^{10}, \rho = 1.2$
2: **while** not converges **do**
3:    Solve $F$ while fixing $A, E$ as,

$$\min_F \; \lambda \|F\|_* + \frac{\mu}{2}\|F - (A + \frac{\Lambda}{\mu})\|_F^2$$

4:    Solve $E$ while fixing $A, F$ as,
5:    **for** $j = 1$ to $m$ **do**
6:       Compute the $j$−th column of $E^*$

$$E^*_{\cdot j} = \frac{P_{\cdot j}}{\|P_{\cdot j}\|_2} \max(\|P_{\cdot j}\|_2 - \frac{1}{\mu}, 0)$$

     where $P = Z^0 - AZ + \frac{1}{\mu}\Omega$
7:    **end for**
8:    Solve $A$ while fixing $E, F$ as,

$$A^* = \left[F + (Z^0 - E + \frac{\Omega}{\mu})Z^T - \frac{\Lambda}{\mu}\right](I + ZZ^T)^{-1}$$

9:    Update multipliers $\Lambda, \Omega, \mu$ as,

$$\Lambda = \Lambda + \mu(A - F)$$
$$\Omega = \Omega + \mu(Z^0 - AZ - E)$$
$$\mu = \min(\rho\mu, \mu_{max})$$

10: **end while**

**Testing step:** For each test image $x$, we first transform it using the proposed CRT $A^*$ as

$$y = A^*x.$$

Then, we identify it by $K$ nearest neighbor (KNN) classifier ($K = 1, 3$) or Sparse Representation Classification (SRC) [Wright *et al.*, 2009] methods. The KNN classifier is performed based on Euclidean distances between $y$ and $Z^0$. For SRC [Wright *et al.*, 2009], we first compute the optimal coefficient $\hat{\alpha}$ as

$$\hat{\alpha} = \mathrm{argmin}_\alpha \|\alpha\|_1 \;\; s.t. \;\; \|y - Z^0\alpha\|_2 \leq \varepsilon$$

Then, we obtain the identity of $y$ as

$$\mathrm{Identity}(y) = \mathrm{argmin}_i \|y - Z^0_i \hat{\alpha}_i\|_2$$

where $Z^0_i$ denote the dataset of the $i$-th class, and $\hat{\alpha}_i$ is the coding coefficient vector associated with the $i$-th class.

## 5 Experiments

To evaluate the recovery ability of CRT method, we first use CRT to recover the true uncorrupted signals from the corrupted images and than conduct classification tasks on the recovered images, as discussed before. Six image datasets are used in the experiments, including three face datasets (AT&T face database, Yale-B [Lee *et al.*, 2005] and CMU-PIE [He *et al.*, 2005]) and three handwritten character datasets (USPS[2], MNIST handwritten digits Database and Binary Alphabet Dataset[3]). Some datasets are resized and Table 1 summarizes the detail setting of these datasets used in the experiments.

Table 3: Dataset descriptions.

| Dataset | # Size | # Dimension | # Class |
|---|---|---|---|
| AT&T | 400 | 644 | 40 |
| Yale-B | 2414 | 896 | 32 |
| CMU PIE | 1500 | 1024 | 30 |
| MNIST | 1500 | 784 | 10 |
| Bin-Alpha | 1014 | 320 | 26 |
| USPS | 1000 | 256 | 10 |

To evaluate the effectiveness and robustness of our method. We conduct classification experiments on both original and corrupted images, respectively. Here, for corrupted images, we randomly add different kinds of corrupted noise including 'block', 'cross' and 'pepper & salt' on each image of the dataset, as shown in Figure 1. The percentage of corruption is about $10\%$ of image size. Then, we test different methods on these corrupted datasets. All experiments are performed with five-fold cross validation strategy, i.e., all data sets are randomly splitted into five equal subsets, iteratively pick four subsets for training and the remaining one subset for the classification testing, then the classification performances are averaged over the five loops. For comparison, we compared our CRT model with original raw data and some other popular data representation and recovery methods including standard Principal Component Analysis (PCA) [Duda *et al.*, 2001], Locality Preserving Projection (LPP) [He *et al.*, 2005], Neighborhood Persevering Embedding (NPE) [He *et al.*, ICCV 2005] and $\ell_1$ norm PCA (L1PCA) [Ke and Kanade, 2005]. All of these methods can learn an optimal projection/transformation from training data and do projection or representation for the test data. We use the K-nearest neighbor (K-NN with K = 1, 3) classifier and Sparse Representation Classification (SRC)[Wright *et al.*, 2009] to evaluate the quality of various transformed representations. We set the regularization parameter $\lambda$ in CRT to 0.12, 0.16 and 0.2, respectively. For standard PCA, LPP, NPE and L1PCA, we set the dimension $d$ to 50 and 100, respectively.

Table 2, 3 show the comparison results using KNN classification on occluded and original image datasets, respectively. From Table 2, we can observe that (1) traditional representation methods LPP and NPE generally break down in presence of corruption noise on face image datasets. (2) L1PCA generally performs better than standard PCA, LPP and NPE methods, indicating the robustness of L1PCA representation method. (3) Our CRT representation obviously outperforms other representation methods. This clearly demonstrates that the proposed CRT can correctly recover the occluded/noise images and thus obviously improves the KNN classification results. Table 3 also lists the results on original image datasets. One can note that our CRT representation

---

[2]http://www.cs.nyu.edu/ roweis/data.html
[3]http://olivier.chapelle.cc/ssl-book/benchmarks.html

Table 1: Comparison of classification results using KNN classification method on occluded image datasets

| Datasets | | Original | PCA | | LPP | | NPE | | L1PCA | | CRT | | |
|---|---|---|---|---|---|---|---|---|---|---|---|---|---|
| | | | $d=50$ | $d=100$ | $d=50$ | $d=100$ | $d=50$ | $d=100$ | $d=50$ | $d=100$ | $\lambda=0.12$ | $\lambda=0.16$ | $\lambda=0.2$ |
| AT&T | 1-NN | 0.625 | 0.570 | 0.558 | 0.370 | 0.353 | 0.400 | 0.435 | 0.730 | 0.618 | 0.858 | 0.858 | 0.863 |
| | 3-NN | 0.700 | 0.588 | 0.588 | 0.385 | 0.355 | 0.403 | 0.465 | 0.740 | 0.630 | 0.825 | 0.830 | 0.838 |
| PIE | 1-NN | 0.557 | 0.497 | 0.478 | 0.253 | 0.287 | 0.286 | 0.295 | 0.626 | 0.647 | 0.667 | 0.665 | 0.661 |
| | 3-NN | 0.550 | 0.497 | 0.475 | 0.247 | 0.287 | 0.289 | 0.302 | 0.643 | 0.661 | 0.649 | 0.645 | 0.645 |
| Yale-B | 1-NN | 0.514 | 0.595 | 0.610 | 0.284 | 0.348 | 0.350 | 0.383 | 0.680 | 0.708 | 0.726 | 0.722 | 0.718 |
| | 3-NN | 0.528 | 0.619 | 0.623 | 0.296 | 0.353 | 0.355 | 0.375 | 0.694 | 0.725 | 0.722 | 0.717 | 0.713 |
| USPS | 1-NN | 0.795 | 0.831 | 0.797 | 0.746 | 0.717 | 0.751 | 0.722 | 0.831 | 0.808 | 0.854 | 0.853 | 0.853 |
| | 3-NN | 0.825 | 0.837 | 0.827 | 0.761 | 0.736 | 0.772 | 0.740 | 0.846 | 0.831 | 0.853 | 0.854 | 0.854 |
| MNIST | 1-NN | 0.735 | 0.749 | 0.722 | 0.251 | 0.272 | 0.290 | 0.298 | 0.770 | 0.770 | 0.784 | 0.786 | 0.786 |
| | 3-NN | 0.750 | 0.759 | 0.741 | 0.281 | 0.286 | 0.310 | 0.328 | 0.776 | 0.772 | 0.778 | 0.780 | 0.782 |
| BinAlpha | 1-NN | 0.577 | 0.722 | 0.710 | 0.633 | 0.596 | 0.644 | 0.640 | 0.731 | 0.715 | 0.736 | 0.736 | 0.734 |
| | 3-NN | 0.582 | 0.729 | 0.702 | 0.636 | 0.593 | 0.653 | 0.637 | 0.736 | 0.711 | 0.736 | 0.738 | 0.737 |

Table 2: Comparison of classification results using KNN classification method on original image datasets

| Datasets | | Original | PCA | | LPP | | NPE | | L1PCA | | CRT | | |
|---|---|---|---|---|---|---|---|---|---|---|---|---|---|
| | | | $d=50$ | $d=100$ | $d=50$ | $d=100$ | $d=50$ | $d=100$ | $d=50$ | $d=100$ | $\lambda=0.12$ | $\lambda=0.16$ | $\lambda=0.2$ |
| AT&T | 1-NN | 0.963 | 0.965 | 0.963 | 0.855 | 0.878 | 0.855 | 0.893 | 0.963 | 0.970 | 0.978 | 0.980 | 0.980 |
| | 3-NN | 0.925 | 0.933 | 0.925 | 0.833 | 0.858 | 0.853 | 0.878 | 0.933 | 0.945 | 0.948 | 0.948 | 0.950 |
| PIE | 1-NN | 0.913 | 0.901 | 0.902 | 0.736 | 0.807 | 0.793 | 0.825 | 0.905 | 0.919 | 0.921 | 0.922 | 0.922 |
| | 3-NN | 0.910 | 0.891 | 0.902 | 0.715 | 0.787 | 0.795 | 0.827 | 0.896 | 0.903 | 0.914 | 0.913 | 0.915 |
| Yale-B | 1-NN | 0.880 | 0.830 | 0.878 | 0.509 | 0.582 | 0.582 | 0.651 | 0.830 | 0.880 | 0.882 | 0.882 | 0.894 |
| | 3-NN | 0.882 | 0.860 | 0.882 | 0.519 | 0.597 | 0.599 | 0.657 | 0.863 | 0.894 | 0.894 | 0.903 | 0.894 |
| USPS | 1-NN | 0.905 | 0.910 | 0.919 | 0.895 | 0.894 | 0.888 | 0.881 | 0.927 | 0.919 | 0.919 | 0.920 | 0.921 |
| | 3-NN | 0.910 | 0.912 | 0.917 | 0.904 | 0.896 | 0.899 | 0.888 | 0.924 | 0.916 | 0.923 | 0.923 | 0.923 |
| MNIST | 1-NN | 0.880 | 0.902 | 0.903 | 0.814 | 0.808 | 0.880 | 0.890 | 0.810 | 0.785 | 0.913 | 0.913 | 0.916 |
| | 3-NN | 0.845 | 0.903 | 0.902 | 0.812 | 0.804 | 0.903 | 0.902 | 0.824 | 0.793 | 0.916 | 0.925 | 0.916 |
| BinAlpha | 1-NN | 0.703 | 0.769 | 0.762 | 0.725 | 0.714 | 0.733 | 0.734 | 0.782 | 0.784 | 0.790 | 0.795 | 0.790 |
| | 3-NN | 0.697 | 0.778 | 0.772 | 0.739 | 0.714 | 0.748 | 0.729 | 0.783 | 0.786 | 0.785 | 0.794 | 0.794 |

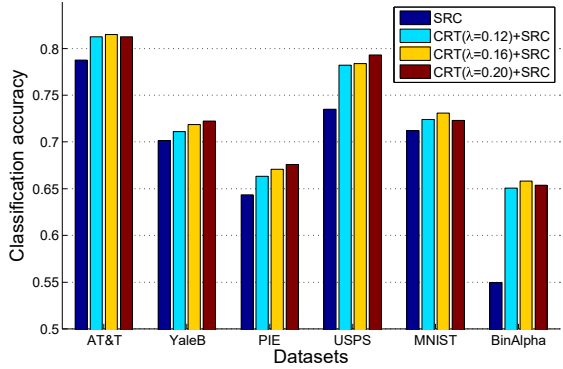
(a) Results on occluded image datasets

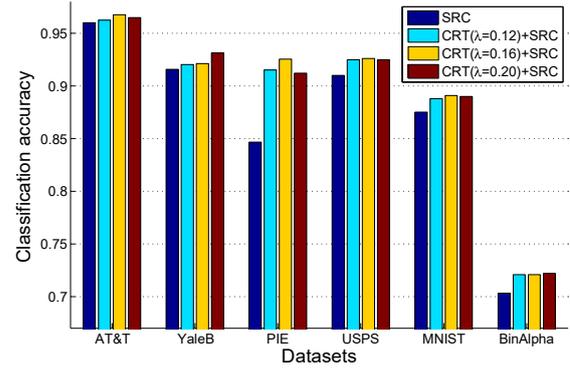
(b) Results on original image datasets

Figure 4: Classification results with CRT+SRC on different datasets

can also outperforms other comparing methods in general. Figure 4 shows the classification results using SRC classification method [Wright *et al.*, 2009]. One can note that (1) SRC obtains reasonable results on occluded image datasets, indicating the robustness of SRC classification methods. (2) CRT obviously improves the SRC classification results on occluded image datasets. This further demonstrates the effectiveness of CRT model in recovering occluded images and thus leads to better SRC classification results.

## 6 Conclusions

In this paper, we introduced a novel Corruption Recovery Transformation (CRT) model which can recover multiple corrupted images using a simple affine transformation. Our CRT model can be efficiently constructed through learning from training data. We formulated CRT construction via a convex learning model (Robust CRT), and derive an effective update algorithm to solve it. As an application, we apply our robust CRT to image classification task. Experimental results demonstrate that the proposed CRT model is effective in recovering noise image data, which obviously improves the classification results.